\documentclass[review]{elsarticle}
\usepackage{lineno,hyperref}
\usepackage{graphicx}
\usepackage{subfigure}
\usepackage{amsmath,amssymb,amsthm}
\usepackage{multirow}
\usepackage{mathrsfs}
\usepackage{booktabs}
\usepackage{amssymb}
\usepackage{accents}
\usepackage{statex}
\usepackage{url}
\usepackage{color}
\usepackage[linesnumbered,ruled,vlined]{algorithm2e}
\SetKwInOut{Notation}{Notations}

\modulolinenumbers[5]
\journal{Journal of \LaTeX\ Templates}

\begin{document}

\begin{frontmatter}

\title{Learning Multi-Tasks with Inconsistent Labels by using Auxiliary Big Task}
\tnotetext[mytitlenote]{Fully documented templates are available in the elsarticle package on \href{http://www.ctan.org/tex-archive/macros/latex/contrib/elsarticle}{CTAN}.}

\author[mymainaddress1,mymainaddress2]{Quan Feng}
\author[mymainaddress1,mymainaddress2]{Songcan Chen\corref{mycorrespondingauthor}}
\cortext[mycorrespondingauthor]{Corresponding author}
\ead{s.chen@nuaa.edu.cn}

\address[mymainaddress1]{College of Computer Science and Technology, Nanjing University of Aeronautics and Astronautics, Nanjing, Jiangsu, 211106, China}
\address[mymainaddress2]{MIIT Key Laboratory of Pattern Analysis and Machine Intelligence, Nanjing University of Aeronautics and Astronautics, Nanjing, Jiangsu, 211106, China}

\begin{abstract}
Multi-task learning is to improve the performance of the model by transferring and exploiting common knowledge among tasks. Existing MTL works mainly focus on the scenario where label sets among multiple tasks (MTs) are usually the same, thus they can be utilized for learning across the tasks. While almost rare works explore the scenario where each task only has a small amount of training samples, and their label sets are just partially overlapped or even not. Learning such MTs is more challenging because of less correlation information available among these tasks. For this, we propose a framework to learn these tasks by jointly leveraging both abundant information from a learnt auxiliary big task with sufficiently many classes to cover those of all these tasks and the information shared among those partially-overlapped tasks. In our implementation of using the same neural network architecture of the learnt auxiliary task to learn individual tasks, the key idea is to utilize available label information to adaptively prune the hidden layer neurons of the auxiliary network to construct corresponding network for each task, while accompanying a joint learning across individual tasks. Our experimental results demonstrate its effectiveness in comparison with the state-of-the-art approaches.
\end{abstract}


\end{frontmatter}

%\linenumbers

\section{Introduction}
Multi-task learning (MTL) is an approach of exploiting and transferring the relevant information among tasks to assist individual tasks obtain better generalization. Over the last few years, it has been proved to be effective in multiple different machine learning fields, such as object detection \cite{chen2018multi}, image segmentation \cite{xu2020multi}, image classification\cite{jiang2016novel}, natural language processing \cite{9053569}, speech recognition \cite{ravanelli2020multi}, drug discovery \cite{zhao2020multi} and so on.

Currently, most existing MTL methods usually assume that learning tasks have the same label sets and use the same model \cite{ruder2017sluice},\cite{2017arXiv170208303B},\cite{2016arXiv161101587H}. Because these label sets contain abundant common knowledge and are transfered to each task to improve the learning performance of the MTL model \cite{2015arXiv150602117L}. However, there are more general situations in the real world, with only a small number of training samples in each task, and when their label sets overlap partially or even unoverlap, there would be less shared information between tasks, so learning such tasks will be more challenging. To meet the challenges, \cite{ma2018modeling} uses a modulation and gating network to automatically adjust the shared characteristics among different tasks for the recommendation system. \cite{2021arXiv210112431F} learns various heterogeneous tasks by sharing similar convolutional kernels among multi-task networks. These methods aim to mine and use as much common knowledge hidden in the current tasks as possible, but for the above-mentioned general scenarios, these methods still leave an improved room in performance.

To achieve the above improvement, we re-focus on the two major issues affecting MTL. Firstly, \emph {how to extract suitable knowledge from different tasks for current multiple tasks.} Since abundant knowledge exists in the nature of multiple tasks that directly affects the joint learning among tasks and useful knowledge for the current tasks can improve the performance of the whole MT model, a key is how to extract this knowledge to avoid notorious negative transfer \cite{2020arXiv200500944W}. For this reason, many methods have been proposed and can be divided into two types: non-deep methods and deep methods: 1) \emph{the non-deep methods build on shallow models to learn the parameters involved}, e.g., \cite{evgeniou2004regularized} extracts useful knowledge between tasks by regularizing a task-coupled kernel function (such as a support vector machine) for the user's prediction of product selection. \cite{honorio2010multi} obtains useful knowledge between tasks by learning the same covariance matrix to predict students' test scores. 2) {\emph{The deep methods learn a shared representation from the individual task networks to improve the performance of the current tasks}. E.g., \cite{Liu2020MultiTaskDF} designs a feature matching  network (i.e., knowledge transfer) to capture shared features in different tasks. \cite{wang2021learning} uses a segmented attention head module to capture useful knowledge between tasks for depth estimation. \cite{2020arXiv200204813G} uses a two-level graph neural network to learn useful knowledge of different tasks to improve the performance of the MTL model. Secondly, \emph {how to design an effective MTL sharing mechanism.} An effective sharing mechanism can increase the predictive performance of the MTL model by using useful knowledge between related tasks \cite{2020arXiv200413379V}. Inspired by this motivation, many classic MTs sharing mechanisms have been designed. According to whether the task's feature/label spaces are consistent among tasks we can divide these mechanisms into two types: homogeneous task sharing and heterogeneous task sharing, as shown in Table \ref{Table:g}. The homogeneous task sharing mechanisms can further be subdivided into 1) \emph{hard sharing} based: the implementation of this type of methods assumes that all tasks share knowledge in the same hidden space. For example, \cite{shen2020deep} connects the aggregated features of specific layers between tasks for semantic segmentation and depth prediction of images. 2) \emph{Soft sharing} based: the implementation of this type of methods assumes that all task models and parameters are independent, and the distance between model parameters is regularized to obtain similar parameters for joint learning. E.g., \cite{2020Identifying} uses the attention mechanism to share parameters in specific layers between different tasks to identify symptoms of depression. 3) \emph{Mixed sharing} based: the implementation of this type of methods uses a special task strategy to select the layer of the multi-task network model can perform shared learning. Typically, \cite{2019arXiv191112423S} uses a specific task strategy to mix these common features with the current tasks for image semantic and normal segmentation. The heterogeneous task sharing mechanism can likewise be further subdivided into 1) \emph{Sparse sharing} based: the implementation of this type of methods is to form a sub-network appropriate for individual tasks from an overparameterized base network, and to extract the common knowledge from the overlapped parts of the sub-networks through the sparse strategy. For example, \cite{sun2020learning} extracts shared parameters as common knowledge to learn individual tasks by a mask in the overlapping part of the sub-networks. 2) \emph{Gradient sharing} based: the implementation of this type of methods uses some similarities to measure the gradient difference between tasks and calculate the nonnegative weights in these tasks, thereby constructing a shared gradient. For example, \cite{2020arXiv200811643V} constructs a shared gradient to measure the gradient difference among individual tasks by cosine distance to predict hospital mortality. 3) \emph{Hierarchical sharing} based: the implementation of this type of methods performs hierarchical sharing for different overlapping areas between multiple tasks. For example, \cite{sanh2019hierarchical} learns common knowledge from different levels of multiple task networks for natural language processing.

Unfortunately, most of the above works are designed for the scenario where the label sets are the same among tasks, rather than for the scenario where the label sets are partially overlapped or even unoverlapped. A few current works design various learning mechanisms for the latter scenario. However, such methods only capture useful knowledge among tasks, which is still difficult to effectively solve the scenario. For this, we propose a novel multi-task learning framework with the help of a big auxiliary deep network (DAMTL), whose intention is to use the auxiliary task(s) with abundant knowledge to assist learning given multiple tasks with partial, even unoverlapping label sets. Our framework is shown in Fig.\ref{Fig:1}. Based on it, we first pre-train a big overparameterized auxiliary network which contains the class label information in all individual tasks; Next, we use a set of soft masks to selectively prune the neurons in the convolutional layers of this network to yield the corresponding network for each learning task. Finally, we jointly train all individual networks in an end-to-end manner.
\begin{figure}[ht]
\centering
\includegraphics[scale=0.9]{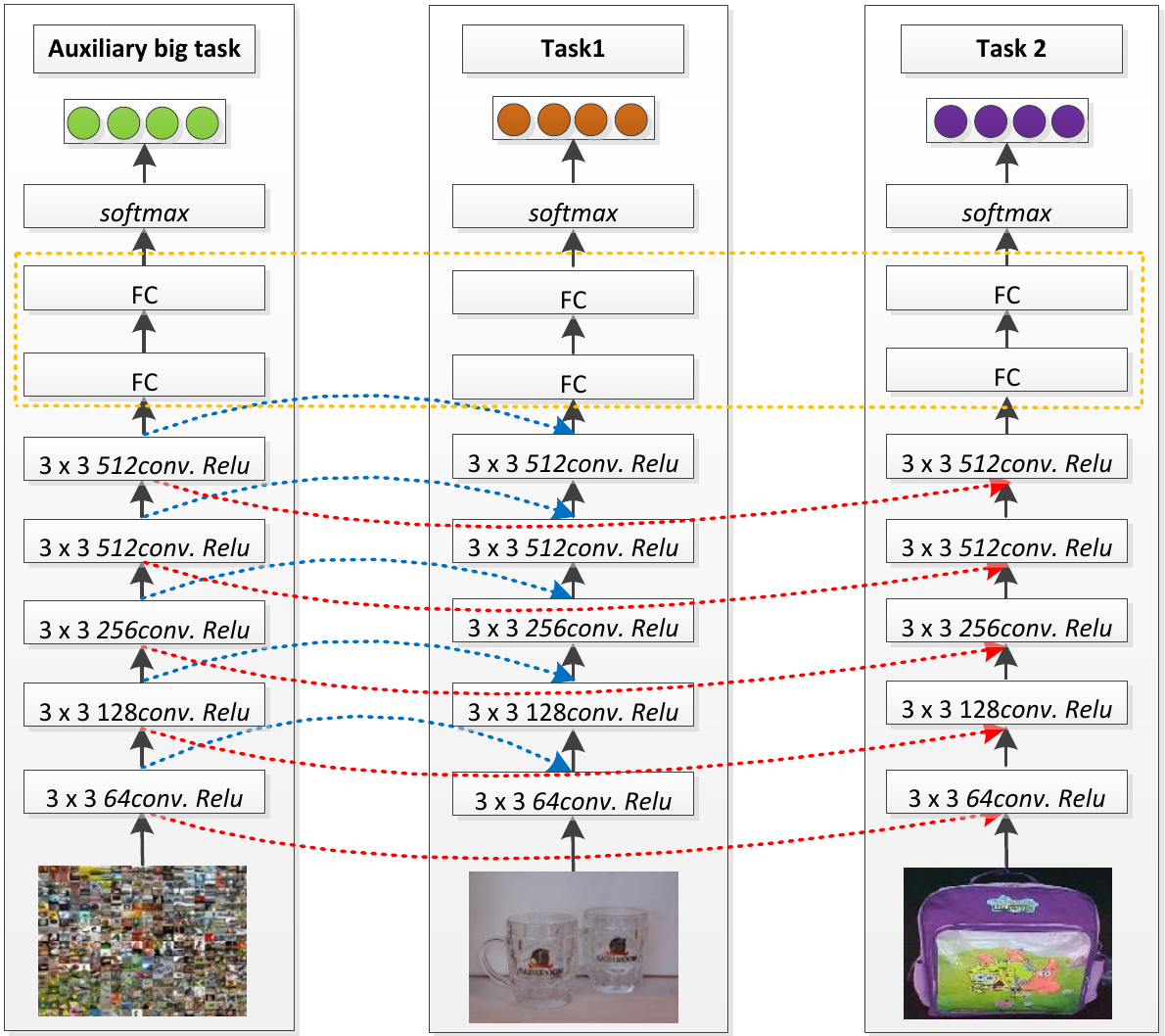}
\caption{Our proposed DAMTL networks. The network consists of three identical independent task networks. The left side is the auxiliary big task, and the rest are individual tasks; {\color{red}$\dashrightarrow$} and {\color{blue}$\dashrightarrow$} indicate the direction in which the knowledge of the auxiliary task is transferred to different tasks, and the yellow dotted box denotes the alignment layers.} \label{Fig:1}
\end{figure}

In summary, our contribution can be summarized as below:

1) We present DAMTL framework and provide a new way to solve the problem of partial overlap or even unoverlap of label sets in MTL.

2) We design a novel knowledge extracting strategy that uses a set of soft masks to prune neurons in the convolutional layers of the auxiliary task network to extract knowledge for each task learning.

3) We propose a new alignment strategy that alleviates the possible class drift in the knowledge transfer from the auxiliary task to individual tasks in the DMATL network.

4) We conduct experiments on twelve public datasets and compare with the state-of-the-art methods to prove the effectiveness of our method.

\begin{table}
\caption{comparison of various MTL sharing mechanisms.}\label{Table:g}
\begin{tabular}{ccccc}
\toprule  %���ӱ���ͷ������
Sharing mechanism& Homogeneous&Heterogeneous&Supervised&Algorithms\\
\midrule  %���ӱ����к���
Hard sharing& $\checkmark$& $\times$&$\checkmark$&\cite{baxter1997bayesian,ruder2019latent} \\
Soft sharing&$\checkmark$& $\times$&$\checkmark$&\cite{ruder2019latent,Strezoski_2019_ICCV}\\
Mixed sharing& $\checkmark$&$\times$&$\checkmark$&\cite{fernando2017pathnet,pironkov2020hybrid}\\
Sparse sharing& $\checkmark$&$\checkmark$&$\checkmark$& \cite{ cao2017sparse,sun2020learning}\\
Gradient sharing&$\times$&$\checkmark$&$\checkmark$& \cite{2020arXiv200811643V,2020arXiv200509910L}\\
Hierarchical sharing&$\times$&$\checkmark$&$\checkmark$&\cite{Sgaard2016DeepML,7849143}\\
\bottomrule %���ӱ����ײ�����
\multicolumn{5}{l}{Homogeneous task, Heterogeneous task, Supervised learning, Representative algorithms }
\end{tabular}
\end{table}

The rest of this paper is arranged as follows. In Section 2, we briefly review related work in multi-task learning. In Section 3, we introduce the architecture of DAMTL, give the definition and some related theoretical application analysis. In the experimental stage of Section 4, we present image classification results on benchmark data sets. Finally, we conclude in Section 5. The code is available at {\color{blue} \emph {http://parnec.nuaa.edu.cn/3021/list.htm}}.

\section{Related Work}
MTL has good performance in many applications, especially in the field of computer vision, so it has attracted a lot of attention in recent years. In this section, we briefly review the related works of MTL based on shared task features and MTL based on shared model parameters. Our work follows the latter research line.

\subsection{MTL based on shared task features}
The methods of this class usually assume that a common feature representation can be learned from individual tasks. According to the implementation manners, they likewise can roughly be divided into three sub-types:

1) \emph{Selective sharing of task features:} for the tasks in the same subspace, they realize sharing by specifically regularizing the features among tasks. Typically, \cite{xue2020weighted} uses the $\ell_2$ norm to regularize the task weight matrices to extract shared features for the test score prediction of most school students. \cite{zhang2019multi} uses the $\ell_{1,2}$ norm to regularize the weight matrices to extract shared features between tasks for learning multi-tasks with different feature dimensions. \cite{shao2020hypergraph} uses $\ell_{2,1}$ norm to regularize the weight matrices of various modal tasks to jointly select common features for multi-modal classification of Alzheimer's disease.

2) \emph{Priori knowledge sharing of tasks:} for the tasks defined in the same subspace, they use the same prior knowledge among tasks to realize sharing. Typically, \cite{2018Multi} embeds prior knowledge (i.e., pathological images with different magnification belong to the same subclass) into the feature extraction process among different tasks to verify the relationship between tasks and pathological image categories for fine-grained classification and pathophysiological image classification. \cite{zheng2019metadata} uses a kind of meta data (i.e., contextual attributes) as a priori knowledge to capture the relationship between different tasks for multiple tasks clustering. \cite{9098703} uses the same subclass of the gland area as the prior information in the convolutional neural network to guide the network inference for pathological colon image analysis.

3) \emph{Transformation sharing of task features:} for the tasks represented in the same subspace, they realize sharing by performing the nonlinear transformations of the original feature representation among tasks. Typically, \cite{misra2016cross} uses a set of non-linearly transformed feature sharing units for image semantic segmentation and normal estimation. \cite{duan2020unsupervised} uses the feature adapter to learn the non-linear transformation of the tasks features to automatically evaluate the child's speech ability.

\subsection {MTL based on shared model parameters}
The methods of this class usually associate different tasks with their partial model parameters or weights to realize sharing. According to their learning manners used, they can roughly be divided into three sub-types:

1) \emph{Weighted sharing of weight matrices:} for the tasks represented in the same subspace, they realize sharing by weightedly combining a set of weight matrices among tasks. Typically, \cite{2017arXiv170400514A} weights the weight matrices among tasks for boundary classification of keywords. \cite{rai2010infinite} partitions the weight matrices among tasks into common and private parts, then weights the common part for multi-label classification. \cite{Zhou_2020_CVPR} weights the weight matrices at the same spatial position in the pictures and transfers them to each task for image depth estimation, segmentation, and surface normal prediction.

2) \emph{Common factor sharing via decomposing individual weight matrices:} for the weight matrix of each task model, they decompose these matrices into private and common parts, where the common part is used for sharing. Typically, \cite{article} decomposes the weight matrices of multiple task models into common and private parts, and further uses the common part for visual target tracking. \cite{8944708} sparsely decomposes the parameter tensor of the prediction model into multiple parameter matrices, and linearly combines the corresponding parameter matrices into a set of base matrices for sharing. \cite{2020Predicting} decomposes a collective matrix of drug-disease correlations to share the correlation matrix between them for drug discovery.

3) \emph{Low-rank structure sharing of model weight matrices:} for the tasks represented in the same subspace, they capture the low-rank structure of the weight matrix among tasks by specifically regularizing to realize sharing. Typically, \cite{2020arXiv200204799Z} uses feature tensor flattening of different tasks (i.e., a convex combination of matrix trace norms) to capture its low-rank structure for multi-task learning. \cite{chen2020template} uses a set of low-rank matrices to capture the potential relationships between multiple tasks for Parkinson's disease diagnosis. \cite{8694882} uses a set of low-rank matrices constrained by the nuclear norm for target detection in hyper-spectral images.

\section{Our Method}
Most of the previously mentioned methods extract useful knowledge among tasks to make predictions. However, these methods are difficult to be further improved when faced with the partially overlapped, even unoverlapped labels among tasks. To overcome this difficulty, we try to leverage a big auxiliary task with abundant labels and class information to assist learning these tasks with limited data. As shown in Fig.\ref{Fig:3}, our method mainly consists of three steps: 1) pre-training a big overparameterized auxiliary network, 2) selectively extracting the corresponding specific weight parameters for individual tasks from the auxiliary network, 3) transferring the weight parameters to these individual tasks to assist them learning. Specifically, Section 3.1 formally defines the problem. Section 3.2 details the proposed method, which extracts knowledge from the pre-trained auxiliary network through a soft making matrix, and then transfers them into individual tasks to form the corresponding networks. Finally, the whole DAMTL network is formulated.
\begin{figure}
\centering
\includegraphics[scale=0.5]{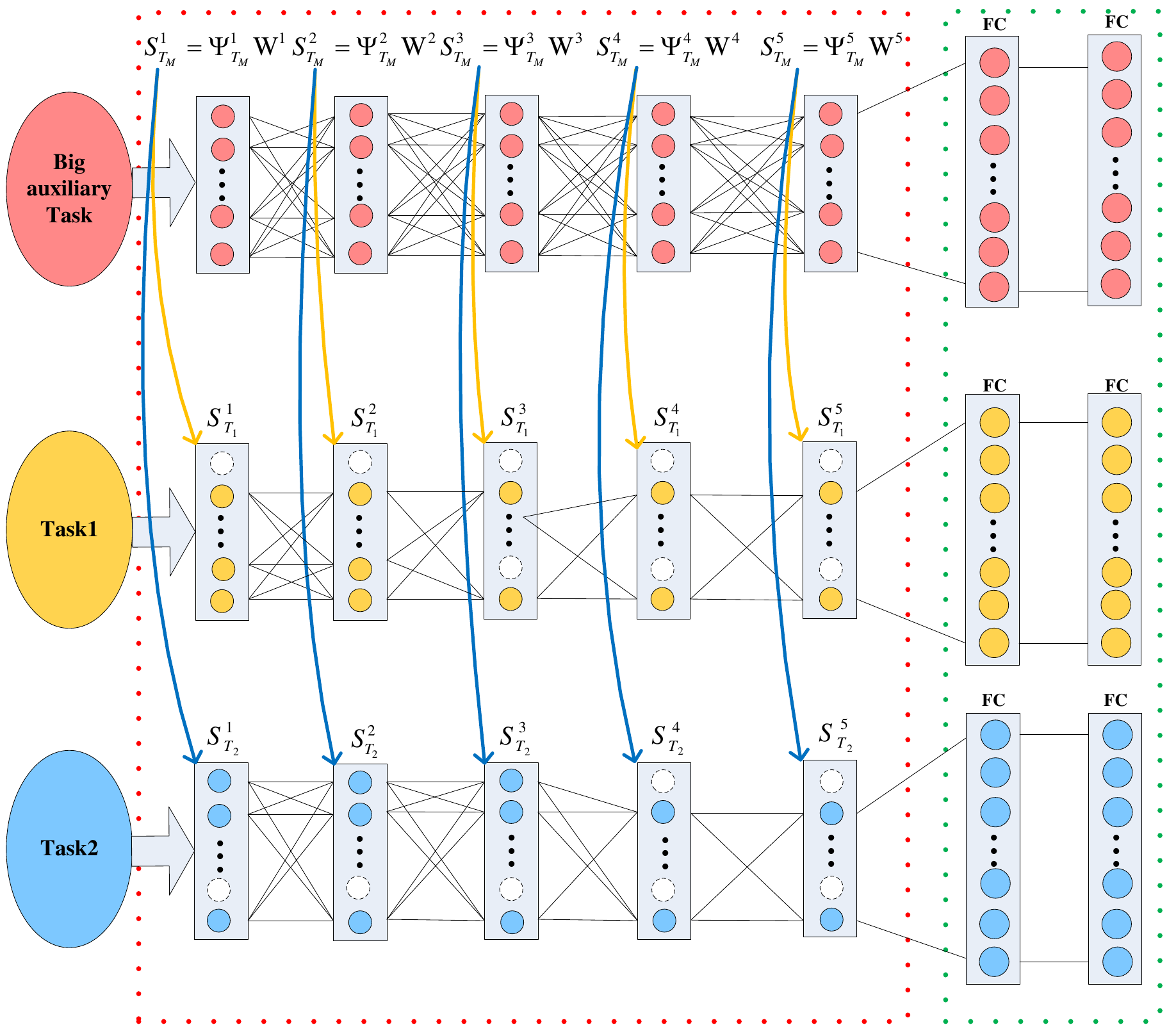}
\caption{Detailed framework of the DMTAL network. The top of the framework is the auxiliary task network, and the rest are different task networks. All the filled circles of different colors denote neurons, while dashed circles are neurons that are pruned.  The red and blue dotted boxes are the weight transfer layers and the alignment layers, respectively.} \label{Fig:3}
\label{fig1}
\end{figure}
\subsection{Problem formulation}
Given a big auxiliary task $T_{aux}$ and a dataset $\mathcal{D}_{aux}=\left\{ x_{i}, y_{i} \right\}_{i=1}^{N}$ containing $N$ samples  with $x_{i} \in \mathbb{R}^{d}$ and its associated label $y_{i} \in\left\{1, \ldots, c\right\}$, where ${d}$ and ${c}$ are the numbers of dimensions and classes in the dataset $\mathcal{D}_{aux}$, respectively. Meanwhile we are given $M$ individual tasks $\left\{{T_j} \right\}_{j=1}^{M}$, and corresponding training dataset $\mathcal{D}_{j}=\left\{ x_{k}^{j}, y_{k}^{j} \right\}_{k=1}^{N_{j}}$ with ${N_{j}}$ samples, $x_{k}^{j} \in \mathbb{R}^{d}$ and its associated label $y_{k}^{j} \in\left\{1, \ldots, c_{j}\right\}$, where ${c_{j}}$ is the number of classes in the dataset $\mathcal{D}_{j}$. We assume that the classset  $\mathcal{C}_{T_{aux}}$ of the auxiliary task contains all the individual tasks classes  $\mathcal{C}_{T}$, namely, $\mathcal{C}_{T_{aux}}=\mathcal{C}_{T_1} \cup  \mathcal{C}_{T_2}\cup,...,\mathcal{C}_{T_M}$, where $C_{T_i}$ and $C_{T_j} (i\neq j)$  can be partially overlapped, or even unoverlapped. This makes DAMTL applicable under more general settings than most existing MTL methods.

\subsection{Implementation}
To learn the above tasks mentioned in Section 3.1, we first assume that this big auxiliary task network has $L$ convolutional layers and $H$ fully connected (FC) layers, and pre-train the network to obtain two sets of weights $\mathrm{W}_{T_{a u x}}=\left\{\mathrm{W}_{aux}^{1}, \mathrm{W}_{aux}^{2}, \ldots, \mathrm{W}_{aux}^{L}\right\}$ and $f_{T_{\text {aux }}}=\left\{f_{\text{aux }}^{1}, f_{\text {aux }}^{2}, \ldots, f_{\text {aux }}^{H}\right\}$, where ${\rm{W}}_{aux}^{\emph{l}}$ and $f_{\text {aux }}^{h}$ represent the weights of the $l$-th convolutional layer and the weights of the $h$-th FC layer, respectively. Then, we manage to acquire necessary knowledge selectively from the trained big task to learn individual tasks by soft masking the weights of each corresponding convolutional layer, as shown in Fig \ref{Fig:4}. In what follows, we take the task $T_j$ as a training example. To selectively extract specific knowledge layer-wisely from the big auxiliary task network, we introduce a soft masking matrix $\Psi_{T_{j}}^{l}$ as follows:

\begin{equation} \label{eq:1}
S_{T_{j}}^{l}=\Psi_{T_{j}}^{l}\odot \mathrm{W}_{aux}^{l},
\end{equation}
here $S_{T_{j}}^{l}$ denotes the extracted knowledge from the auxiliary task, which will be transferred to the $l$-th convolutional layer in task $T_j$ , $\Psi_{T_{j}}^{l}$ just takes non-negative value, and $\odot$ denotes the Hardmard product. Then, we use Eq.\eqref{eq:2} to activate the multiplication of $S_{T_{j}}^{l}$ and $F_{T_{j}}^{l-1}$ to realize the knowledge being transferred

\begin{equation}\label{eq:2}
F_{T_{j}}^{l}=\sigma \left(S_{T_{j}}^{l}  F_{T_{j}}^{l-1} +\mathrm{b}_{T_{j}}^{l} \right),
\end{equation}
where $F_{T_{j}}^{l}$ is the feature representation of the $l$-th convolutional layer of the task $T_j$ network, $\sigma$ is the activation function, and $\mathrm{b}_{T_{j}}^{l}$ is the bias vector.

\begin{figure}[ht]
\centering
\includegraphics[scale=0.5]{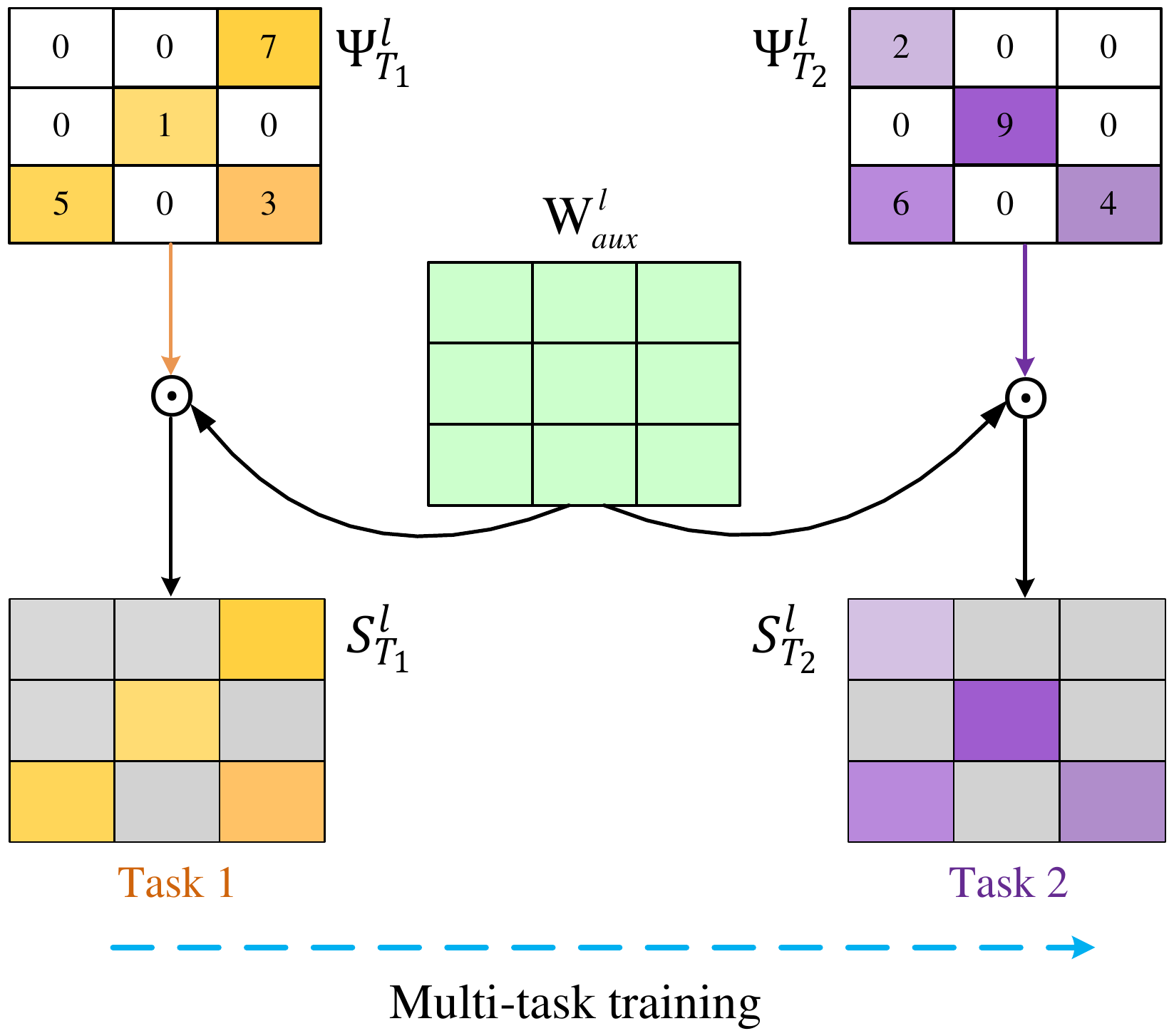}
\caption{Knowledge extraction illustrations. The light green square in the middle represents the weight of the big auxiliary task network in the $l$-th convolutional layer. The squares in the upper left and upper right corners represent the soft masking matrix for different individual tasks, and the squares in the lower left and right corners represent the extracted knowledge.} \label{Fig:4}
\label{fig1}
\end{figure}
Now optimizing the $S_{T_{j}}^{l}$ boils down to optimizing $\Psi_{T_{j}}^{l}$. Specifically, in order to ensure the prediction accuracy of the DAMTL network, we leverage \emph{Conditional Maximum Mean Discrepancy} (CMMD) \cite{long2014transfer} to align the conditional probability distributions between $T_{aux}$ and $T_j$ in the FC layers to alleviate the possible class drift in the knowledge transfer from $T_{aux}$ to the $T_j$, which is expressed as:
\begin{equation} \label{eq:4}
D(T_{\emph{aux }},T_{j})=\sum_{{c_j}=1}^{C_{T_j}}\Vert\frac{1}{k_{c_j}^{T_{j}}} \sum_{k=1}^{k_{c_j}^{T_{j}}} f_\emph{aux}^{h}(O_{k}^{h-1})-\frac{1}{k_{c_j}^{T_{j}}} \sum_{k=1}^{k_{c_j}^{T_{j}}} f_{T_{j}}^{h}(O_{k}^{h-1})\Vert_\mathcal{H}^{2},
\end{equation}
where $(O_{k}^{h-1})$ is the representation of the task $T_j$ in the $(h-1)$-th FC layer, $f_\emph{aux}^{h}$ and $f_{T_{j}}^{h}$ are the weights in the $h$-th FC layer, and $k_{c_j}^{T_{j}}$ is the number of samples of the $c_j$-th class in task $T_j$.

In practice, we use the inputs $x_k $ (${k}=1,2, \ldots,n$) and the labels $y_k$ (${k}=1,2,\ldots,n$) to minimize the following individual task loss
\begin{equation}
\mathcal{L}_{T_j}\left(\hat{y}_{k}^{T_j}, y_{k}^{T_j}\right)=-\sum_{k=1}^{N_{j}} y_k^{T_j}(\log \hat{y}_{k}^{T_j})+ \lambda_1 \sum_{l=1}^{L}\Vert\Psi_{T_{j}}^{l}\Vert_{1}+\lambda_2\sum_{h=1}^{H} D(T_{\emph{aux }},T_{j}),
\end{equation}
where $\hat{y}_{k}^{T_j}$ is the predicted output, $\lambda_1$ and $\lambda_2$ are hyper-parameters, the first term is the cross-entropy loss of task $T_j$, and the second term uses the $\ell_1$ norm to make the soft masking matrix $\Psi_{T_{j}}^{l}$ spare. Finally, we conduct joint training of these tasks with the total loss

\begin{equation} \label{eq:3}
\mathcal{L}(\theta)=\sum_{j=1}^{M} \alpha_{T_j} \sum_{k=1}^{N_{j}} \mathcal{L}_{T_j}\left(\hat{y}_{k}^{T_j}, y_{k}^{T_j}\right),
\end{equation}
where $\alpha_{T_j}$ is the hyper-parameter to be adjusted.

The whole process of the proposed method to solve DAMTL is summarized in Algorithm \eqref{AL:1}.

\begin{algorithm}
        \caption{DAMTL} \label{AL:1}
        \label{alg:opthyperpar}
          \Notation{$\mathrm{W}_{T_{j}}=\left\{\mathrm{W}_{T_j}^{l}\right\}_{l=1}^{L}$ and $\Psi_{T_{j}}=\left\{\Psi_{T_{j}}^{l}\right\}_{l=1}^{L}$ denote the  \\
           \noindent temporary variables and soft masking matrices in the task $T_j$ network, respectively, $f_{T_{j}}=\left\{f_{T_{j}}^{h}\right\}_{h=1}^{H}$
           are the weights of the FC layers in the task $T_j$ network.
          }
          \KwIn{ $D_j$, $\mathrm{W}_{T_{aux}}$, $\lambda_1$, $\lambda_2$}
          \KwOut{$\mathrm{W}_{T_{j}}$, $f_{T_{j}}$, $\Psi_{T_{j}}^{l}$}
                    random initialization $\Psi_{T_{j}}^{l}$, initialize $\mathrm{W}_{T_{j}}$ with $\mathrm{W}_{T_{aux}}$\\
             \Repeat{convergence}{
                    \For{each convolution layer $l$}
                    {
                       $S_{T_{j}}^{l} \leftarrow\left(\Psi_{T_{j}}^{l} \odot \mathrm{W}_{T_j}^{l}\right)$\\
                       $F_{T_{j}}^{l}=\sigma\left(S_{T_{j}}^{l} F_{T_{j}}^{l-1}+b_{T_{j}}^{l}\right)$ \\
                       end for}
                    \For{each fully connected layer $h$}
                    {
                      align the conditional probability distributions between $T_{aux}$ and $T_j$ by Eq.\eqref{eq:4}\\
                    end for}
                    Update the parameters $\mathrm{W}_{T_{j}}^{l}$, $\Psi_{T_{j}}$ and $f_{T_{j}}$ by the back-propagation algorithm
             }
    \end{algorithm}

\section{Experiments}
In this section, we use the VGG network \cite{2014arXiv1409.1556S} as the base-model and the ImageNet dataset as a big auxiliary task to conduct two categories of experiments: 1) the label sets among tasks partially overlap; 2) the label sets among tasks do not overlap.

\subsection{Datasets}
We conduct experiments on the following data sets and divide $70\%$ of the data as training set and the remaining $30\%$ as testing set. The detailed information is shown in Table \ref{Table:1} and Table \ref{Table:2}.

The \textbf{ImageNet Dataset}\footnote{\url{http://image-net.org/download }}
is a computer vision dataset, which is used as an auxiliary task in the experiment. The dataset contains 21,841 categories and 14,197,122 images.

The \textbf{Office-Caltech Dataset}\footnote{\url{https://people.eecs.berkeley.edu/~jhoffman/domainadapt/}}
is divided into two datasets, Office-Caltech10 and Office-Caltech31, and each dataset has 2,533 images, which are composed of three different subsets Dslr, Amazon and Webcam. We randomly select 10 categories in each subset to do experiment.

\begin{table}
\caption{Summarize statistics for datasets where  part of the label sets overlap.}\label{Table:1}
\begin{tabular}{cccc}
\toprule  %���ӱ���ͷ������
Datasets& classes& Size of image&Overlapping classes\\
\midrule  %���ӱ����к���
Caltech-101& 10&$200*300$& 3\\
Caltech-256&10&$371*326$& 3\\
Amazon& 10&$150*900$/$557*28$& 3\\
Webcam& 10&$200*150$/$900*557$& 3\\
Dlsr&10&$200*150$/$900*557$& 3\\
Product& 10&$117*85$/$4384*2686$& 3\\
\bottomrule %���ӱ����ײ�����
\end{tabular}
\end{table}

The \textbf{Office Home Dataset}\footnote{\url{http://hemanthdv.org/OfficeHome-Dataset/}}
is composed of Art, Clipart, Product, Real-World, and each subset covers 15500 images from 65 categories. We also randomly select 10 categories in each subset as the dataset.

The \textbf{Caltech-256 Dataset}\footnote{\url{http://www.vision.caltech.edu/Image_Datasets/Caltech256/}}
is a dataset collected by the California Institute of Technology. It has 256 categories, and each category has more than 80 images and a total of 30,607 images. We also randomly select 10 categories in this data set for experiment.

The \textbf{Tiny ImageNet Dataset}\footnote{\url{https://www.kaggle.com/c/tiny-imagenet}}
is a dataset collected by Stanford University, which is a subset of the ImageNet dataset. The dataset contains 200 categories and a total of 100,000 images, and the size of each image is $3*64*64$. We randomly select 10 categories from the Tiny ImageNet Dataset for experiments.

\begin{table}
\caption{Summarize the statistics of the datasets with unoverlapping label sets.}\label{Table:2}
\begin{tabular}{cccc}
\toprule  %���ӱ���ͷ������
Datasets& classes&Size of image& Overlapping classes\\
\midrule  %���ӱ����к���
Art& 10&$117*85$/$4384*2686$& ---\\
Real World&10&$117*85$/$4384*2686$& ---\\
Caltech-101& 10&$200*300$& ---\\
Webcam&10&$200*150$/$900*557$& ---\\
Amazon& 10&$200*150$/$900*557$& ---\\
Tiny ImagNet& 10&$64*64$& ---\\
\bottomrule %���ӱ����ײ�����
\end{tabular}
\end{table}

\subsection{Comparison Methods}
For evaluation, we use common single-task and multi-tasks network architectures to train each task separately/jointly, and its experimental results serve as our single-task and multi-tasks baseline. Simultaneously, we compare our proposed method with the following MTL methods:

\textbf{\textbf{Single task}}\footnote{\url{https://github.com/machrisaa/tensorflow-vgg}} \cite{2014arXiv1409.1556S}: This method uses a single VGG network to learn the predictive model for each independent task.

\textbf{\textbf{Multi-task}}\footnote{\url{https://github.com/luntai/VGG16_Keras_TensorFlow}}: This method uses multiple identical VGG networks to jointly learn a multi-task prediction model.

\textbf{\textbf{Cross-Stitch}}\footnote{\url{https://github.com/helloyide/Cross-stitch-Networks-for-Multi-task-Learning}} \cite{misra2016cross-stitch}: This method uses a cross stitch unit to learn the common features in two network feature layers for learning multiple tasks.

\textbf{\textbf{NDDR-CNN}}\footnote{\url{https://github.com/ethanygao/NDDR-CNN}} \cite{gao2019nddr-cnn:}: This method uses the NDDR module to automatically integrate the features of each sub-network layer for MTL.

\textbf{\textbf{MTAL}}\footnote{\url{http://parnec.nuaa.edu.cn/3021/list.htm}}\cite{2021arXiv210112431F}: This method leverages the similarity between convolution kernels to capture common knowledge among multi-network for joint learning of various tasks.

\textbf{\textbf{DAMTL}}\footnote{\url{http://parnec.nuaa.edu.cn/3021/list.htm}}: This method uses the abound knowledge of a lager auxiliary task to help joint learning of multiple tasks.

\subsection{Hyper-Parameter Tuning}
In the contrasted deep neural network methods, we adjust the hidden units, learning rate, and the number of training steps in each layer according to the parameter settings of the corresponding reference. In DAMTL, we adjust the hyper-parameters in the same way. Specifically, for all experiments, we set the learning rate $\eta$ to 0.01, $\alpha_{T_j}$ to 0.01, $\lambda_1$ and $\lambda_2$ are 0.9. In addition, in DAMTL network training, we select the rectified linear unit (ReLU) function as the activation function $\sigma$. All the deep learning models are implemented by Tensorflow.

\subsection{Results of Model Performance:}
\setlength{\parindent}{2em}
We conduct experiments on the above-mentioned datasets and compare our methods with the state-of-the-arts (SOTAs), while analyzing the experimental results.

First, the performance of the DAMTL method shown in Table \ref{Table:3} and Table \ref{Table:4} is significantly better than other methods. In the PO-2 group of experiments, the
DAMTL method is better than the MTAL methods.

\begin{table}
\centering
\caption{The performance comparison of various methods in the experiment of partially overlapping label sets between tasks. Among them, the bold numbers are the best classification results, and the underlined numbers are the sub-optimal classification results.}\label{Table:3}
\scriptsize
\label{my-label}
\begin{tabular}{c|c|c|c|c|c|c}
\hline
\multicolumn{1}{c|}{\multirow{2}{*}{Methods}} & \multicolumn{2}{c|}{PO-1} & \multicolumn{2}{c|}{PO-2} & \multicolumn{2}{c}{PO-3} \\ \cline{2-7}
\multicolumn{1}{c|}{}&Caltech-101&Caltech-256&Amazon&Webcam&Dlsr&Product\\ \hline
Single-Task&$0.76\pm 0.037$&$0.45\pm0.037$&$0.74\pm0.036$&$0.63\pm0.038$&$0.77\pm0.031$&$0.63\pm0.035$\\
Multi-task&$0.76\pm0.050$&$0.53\pm0.055$&$0.80\pm0.046$&$0.69\pm0.049$&$0.80\pm0.046$&$\underline{0.68\pm0.049}$\\
Cross-Stich&$0.75\pm0.050$&$0.53\pm0.059$&$0.80\pm0.046$&$0.67\pm0.054$&$0.79\pm0.042$&$0.65\pm0.056$ \\
NDDR-CNN&$0.75\pm0.054$&$0.51\pm0.055$&$0.79\pm0.042$&$0.67\pm0.044$&$0.75\pm0.049$&$0.67\pm0.050$ \\
MTAL&$\underline{0.78\pm0.054}$&$\underline{0.53\pm0.051}$&$\mathbf{0.81\pm0.042}$&$\underline{0.69\pm0.048}$&$\underline{0.80\pm0.038}$&$0.66\pm0.055$\\
$\mathbf{DAMTL}$&$\mathbf{0.80\pm0.050}$&$\mathbf{0.54\pm0.059}$&$\underline{0.80\pm0.044}$&$\mathbf{0.70\pm0.048}$&$\mathbf{0.84\pm0.042}$&$\mathbf{0.74\pm0.052}$\\ \hline
\end{tabular}
\end{table}
Secondly, Table \ref{Table:3} shows that most of the MTL methods are better than the single-task learning method, which demonstrate that using the relationship between tasks to capture their useful information for interaction can promote the effectiveness of the joint learning of multiple tasks. In addition, we find that different  multi-task learning methods have different performance results, which are caused by the differences among tasks. For example, the experimental results of the dataset Caltech-101 in PO-1 show that the Cross-Stich method and the NDDR-CNN method are lower than the accuracy of the single-task learning method.

\begin{table}
\centering
\caption{Performance comparison of various methods in unoverlapping experiments of label sets between tasks. Among them, the bold numbers are the best
classification results, and the underlined numbers are the sub-optimal classification results.}\label{Table:4}
\scriptsize
\label{my-label}
\begin{tabular}{c|c|c|c|c|c|c}
\hline
\multicolumn{1}{c|}{\multirow{2}{*}{Methods}} & \multicolumn{2}{c|}{NO-1} & \multicolumn{2}{c|}{NO-2} & \multicolumn{2}{c}{NO-3} \\ \cline{2-7}
\multicolumn{1}{c|}{}&Art&Real World& Caltech-101 &Webcam & Amazon &T-ImagNet\\ \hline
Single-Task&$0.60\pm 0.039$&$0.51\pm 0.043$&$0.76\pm 0.037$&$0.63\pm 0.038$&$0.77\pm 0.031$&$0.51\pm 0.048$ \\
Multi-task&$0.62\pm 0.055$&$\underline{0.53\pm 0.054}$&$0.78\pm 0.050$&$0.68\pm 0.050$&$0.78\pm 0.048$&$0.53\pm 0.058$\\
Cross-Stich&$\underline{0.63\pm 0.057}$&$0.51\pm 0.056$&$\underline{0.78\pm 0.048}$&$0.67\pm 0.050$&$0.79\pm 0.046$& $0.52\pm 0.062$ \\
NDDR-CNN&$0.59\pm 0.056$&$0.52\pm 0.056$&$0.76\pm 0.051$&$0.66\pm 0.050$&$0.79\pm 0.044$&$0.46\pm 0.461$ \\
MTAL&$0.61\pm 0.056$&$0.52\pm 0.058$&$0.73\pm 0.046$&$\underline{0.69\pm 0.044}$&$\underline{0.80\pm 0.042}$&$\underline{0.54\pm 0.067}$ \\
$\mathbf{DAMTL}$&$\mathbf{0.67\pm 0.050}$&$\mathbf{0.6\pm 0.047}$&$\mathbf{0.80\pm 0.050}$&$\mathbf{0.72\pm 0.044}$&$\mathbf{0.81\pm 0.047}$&$\mathbf{0.54\pm 0.065}$ \\ \hline
\end{tabular}
\end{table}

Finally, the performance of DAMTL is better than that other methods in all datasets, which shows that extracting and transferring features from a big auxiliary task (which contains the class label information in all individual tasks) can help the joint learning of multiple tasks with partially overlap or even unoverlapping label sets and further improve DAMTL performance.

Also, Fig.\ref{Fig:5} shows the performance comparison of the mean and standard deviation of the classification accuracy of various methods in the two categorical experiments. We observe that the overall performance of the DAMTL method is better than that other methods. The above experimental results are consistent with our theoretical analysis.

\begin{figure} \centering
\subfigure[H] { \label{fig:a}
\includegraphics[width=0.47\columnwidth]{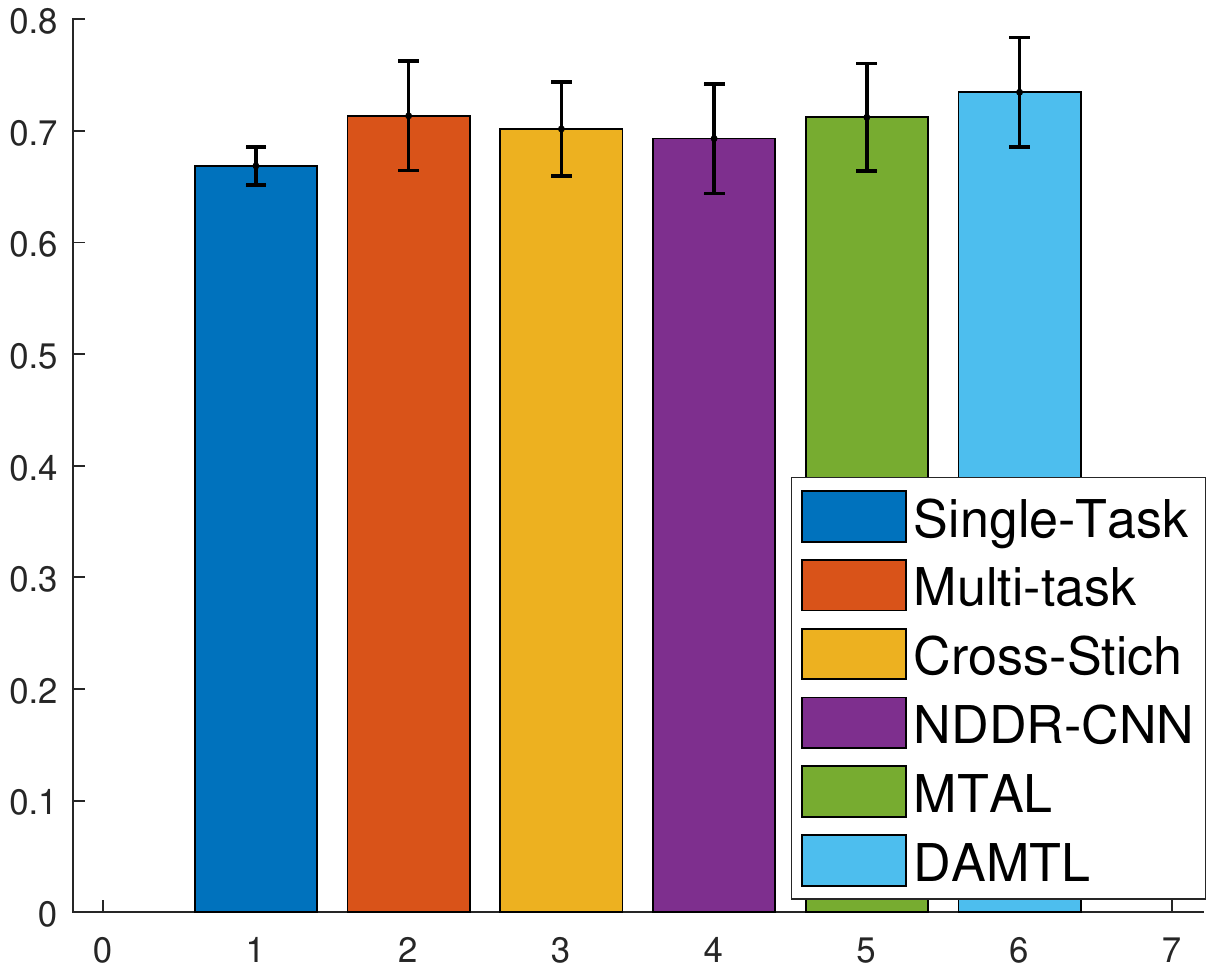}
}
\subfigure[] { \label{fig:b}
\includegraphics[width=0.449\columnwidth]{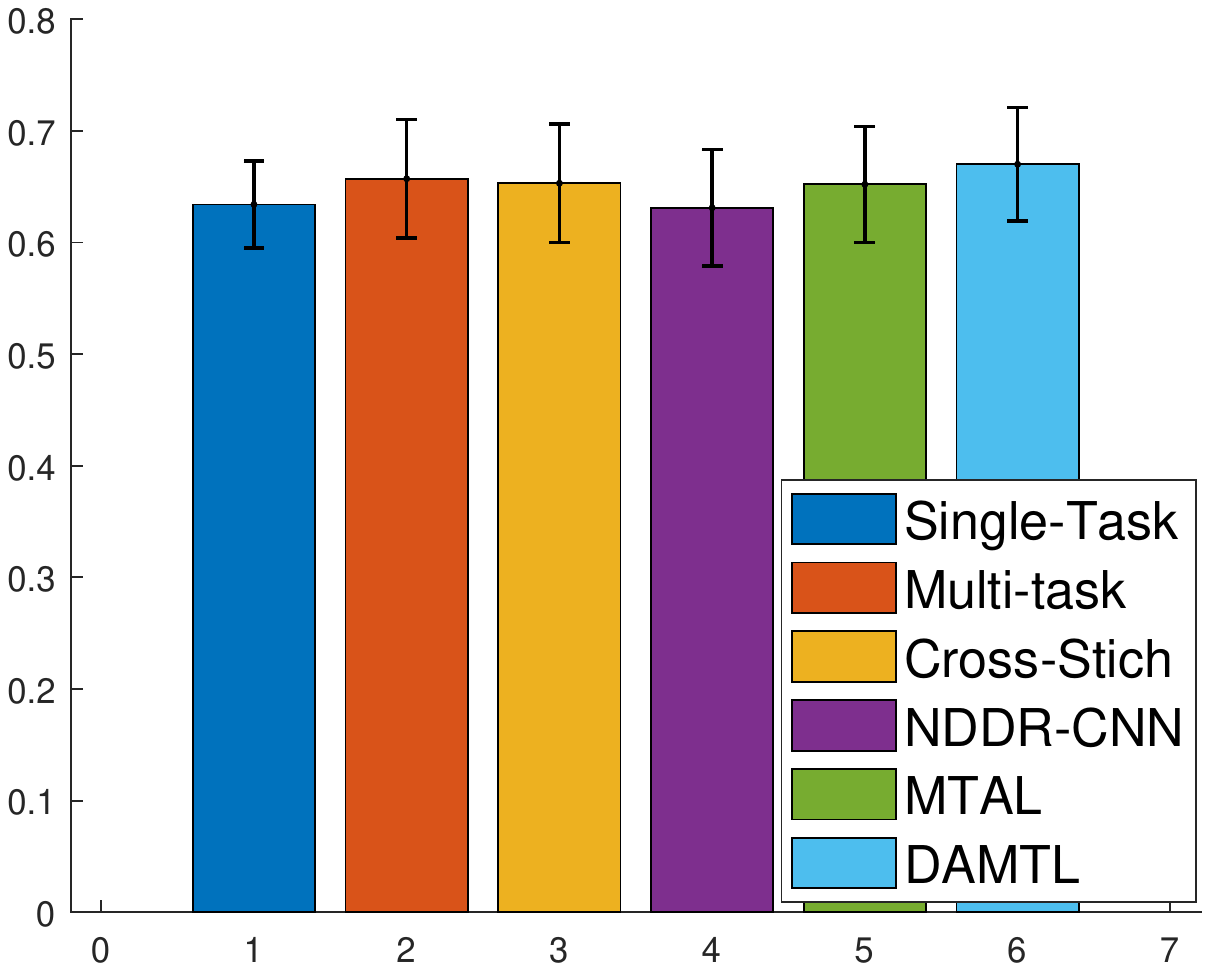}
}
\caption{Performance comparison of various methods on mean and mean square error.  Sub-Fig.(a) shows that the performance of various methods in the mean and mean square error of the scenario where the label sets between tasks are partially overlapped is various.  Sub-Fig.(b) shows that the performance of various methods in the mean value and mean square error of the scenario where the label sets between tasks are completely unoverlapping is various.}\label{Fig:5}
\label{fig}
\end{figure}

\subsection{Time cost comparison}
The result is shown in Fig.\ref{Fig:f} (a), we observe that when the label sets among tasks partially overlap, although the single-task method takes the shortest time, it does not use shared information, so the accuracy is the lowest. The NDDR-CNN method takes the longest time, but the accuracy is the second lowest (even worse than the multi-task benchmark). This indicates that the differences among tasks lead to adversarial interference in the learning process of this method. In addition, the multitasking, Cross-Stich, NDDR-CNN and MTAL methods are comparable in time cost, and our DAMTL method has the highest accuracy. As shown in Fig.\ref{Fig:f} (b), we observe that when the label sets among tasks do not overlap, the single-task method takes the shortest time, but its accuracy is the lowest. MTAL uses the convolution kernel sharing technology, thus spending the second shortest time. In conclusion, our method has relatively low time cost and the highest accuracy in scenarios where the task label sets are partially overlapped or even unoverlapped.

\begin{figure}  \centering
\subfigure[] { \label{fig:a}
\includegraphics[width=0.408\columnwidth]{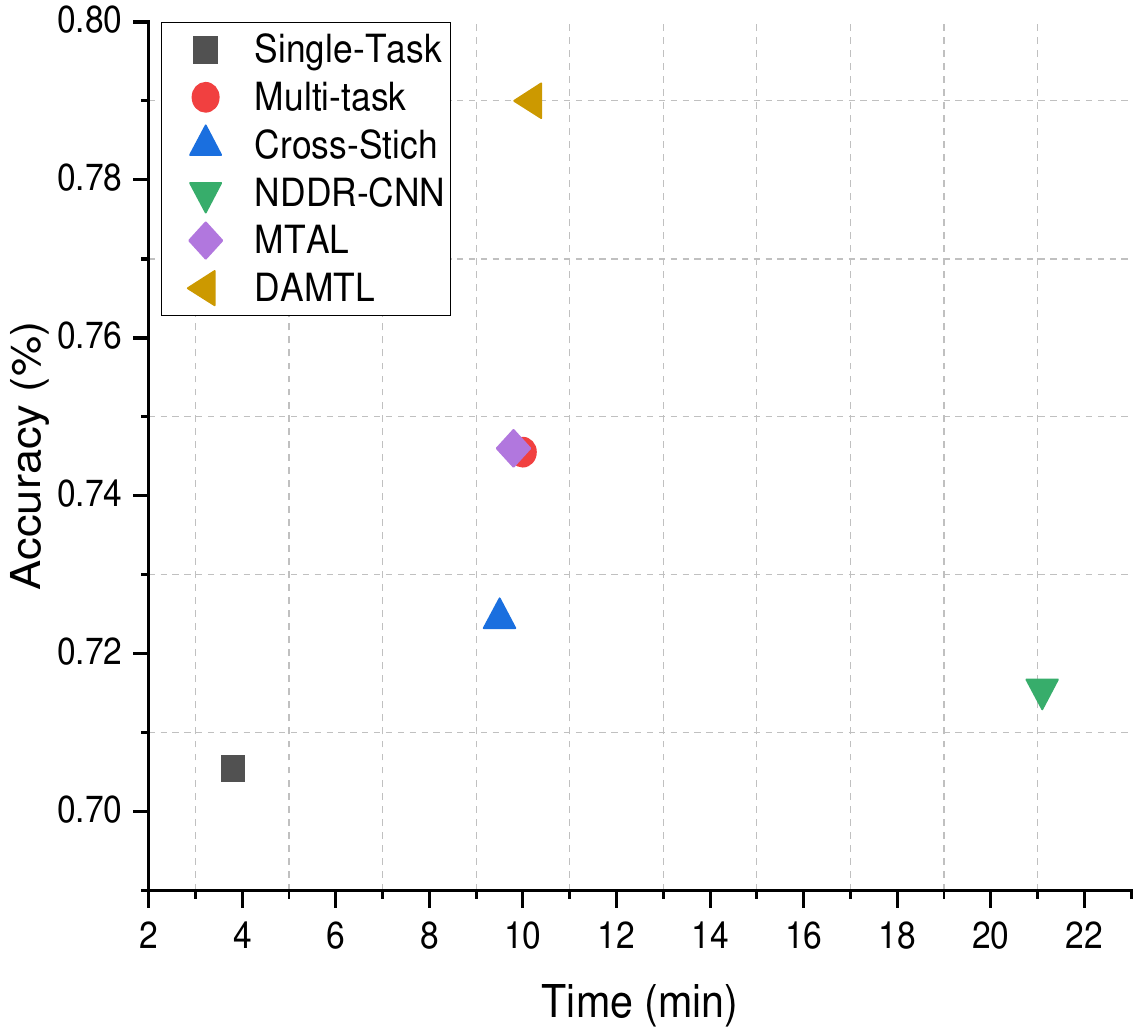}
}
\subfigure[] { \label{fig:b}
\includegraphics[width=0.42\columnwidth]{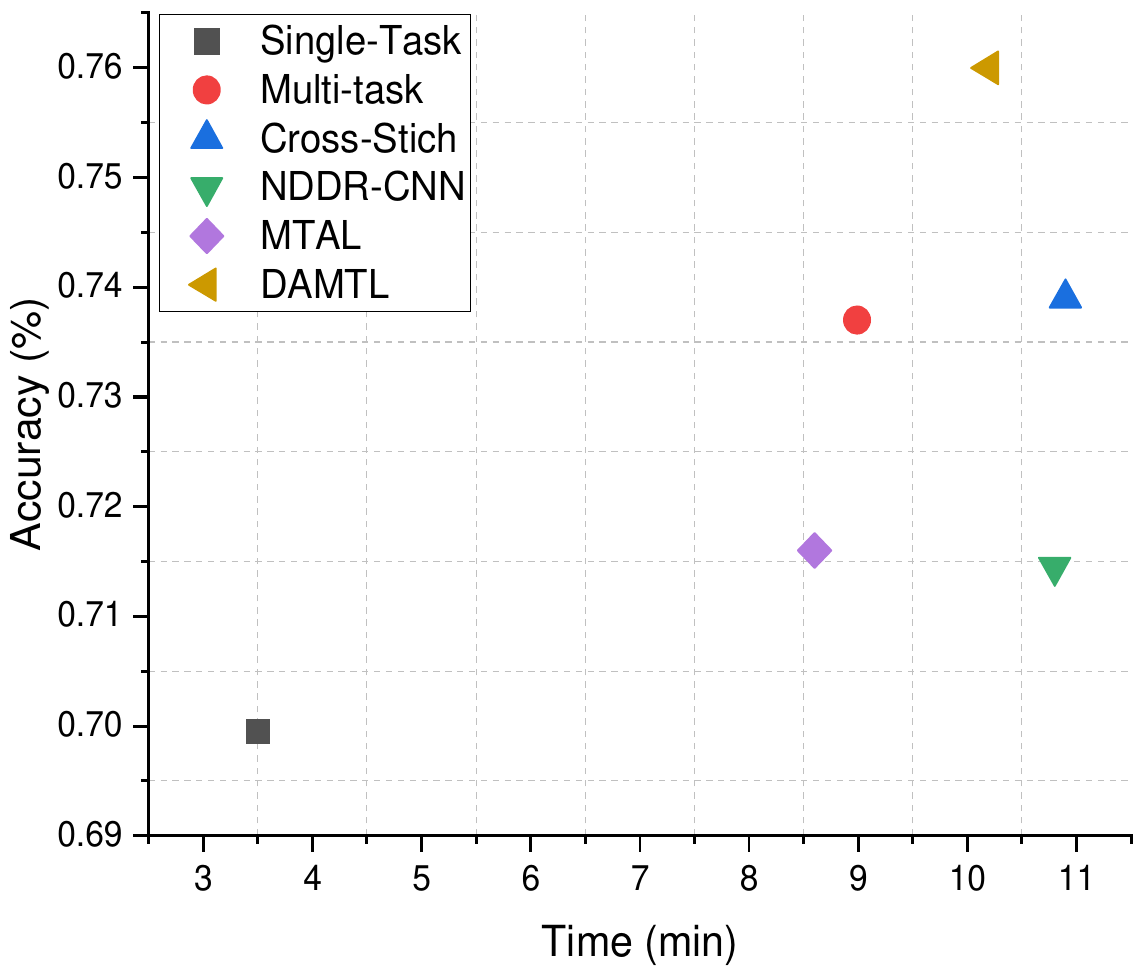}
}
\caption{Comparison of the time cost (in minutes) for MTL  on partially overlapping and unoverlapping label sets.
}\label{Fig:f}
\label{fig}
\end{figure}

\subsection{Ablation Study}
In this section, we demonstrate that the addition of Eq.\eqref{eq:4} to DAMTL through ablation analysis is very important to achieve state-of-the-art performance. We first study the performance of multiple tasks with overlapping partial label sets without using Eq.\eqref{eq:4}. The performance of DAMTL network on Caltech-101, Caltech-256, Amazon, Webcam, Dlsr, and Product is shown on Table \ref{Table:6}. Obviously, when the rest of the data set except Product, the performance of DAMTL is reduced.
\begin{table}
\centering
\caption{Results of ablation study when task label sets partially overlap. $DAMTL_{Fa}$ represents the DAMTL network without feature alignment.}\label{Table:6}
\scriptsize
\label{my-label}
\begin{tabular}{c|c|c|c|c|c|c}
\hline
\multicolumn{1}{c|}{\multirow{2}{*}{Methods}} & \multicolumn{2}{c|}{PO-1} & \multicolumn{2}{c|}{PO-2} & \multicolumn{2}{c}{PO-3} \\ \cline{2-7}
\multicolumn{1}{c|}{}&Caltech-101&Caltech-256&Amazon&Webcam&Dlsr&Product\\ \hline
DAMTL&$0.80\pm0.050$&$0.54\pm0.059$&$0.80\pm0.044$&$0.70\pm0.048$&$0.84\pm0.042$&$0.74\pm0.052$\\
$DAMTL_{Fa}$&$0.75\pm0.053$&$0.50\pm0.057$&$0.79\pm0.046$&$0.65\pm0.054$&$0.63\pm0.049$&$0.76\pm0.052$\\ \hline
\end{tabular}
\end{table}
Then, we delete the Eq.\eqref{eq:4} part for unoverlapping label sets between tasks. The performance of DAMTL network on Art, Real World, Caltech-101, Webcam, Amazon, and T-ImagNet is shown on Table \ref{Table:7}. The performance of DAMT on all data sets is significantly degraded. Finally, we use Eq.\eqref{eq:4} DAMTL network through ablation analysis which can effectively improve performance.
\begin{table}
\centering
\caption{Results of ablation study when task label sets do not overlap. $DAMTL_{Fa}$ represents the DAMTL network without feature alignment.}\label{Table:7}
\scriptsize
\label{my-label}
\begin{tabular}{c|c|c|c|c|c|c}
\hline
\multicolumn{1}{c|}{\multirow{2}{*}{Methods}} & \multicolumn{2}{c|}{NO-1} & \multicolumn{2}{c|}{NO-2} & \multicolumn{2}{c}{NO-3} \\ \cline{2-7}
\multicolumn{1}{c|}{}&Art&Real World& Caltech-101 &Webcam & Amazon &T-ImagNet\\ \hline
DAMTL&$0.67\pm 0.050$&$0.6\pm 0.047$&$0.80\pm 0.050$&$0.72\pm 0.044$&$0.81\pm 0.047$&$0.54\pm 0.065$ \\
$DAMTL_{Fa}$&$0.61\pm 0.048$&$0.48\pm 0.048$&$0.75\pm 0.044$&$0.64\pm 0.050$&$0.79\pm 0.041$&$0.50\pm 0.045$ \\ \hline
\end{tabular}
\end{table}

\subsection{Model convergence analysis}
In this section, we analyze the convergence of our proposed method on two benchmarks, i.e., Caltech-101 and Caltech-256 and the values of the objective function \eqref{eq:3} with respect to iterations on the two datasets are shown in Fig.\ref{Fig:6} (a) and (b) respectively. From Fig.\ref{Fig:6} (a), we can see that the loss curves on the data sets Caltech-101 and Caltech-256 tend to converge after about 50 iterations. In Fig.\ref{Fig:6} (b), the loss curves on the data sets Art and Real-world stabilize after about 70 and 50 iterations, respectively.

\begin{figure} [h]
\subfigure[] { \label{fig:a}
\includegraphics[width=0.41\columnwidth]{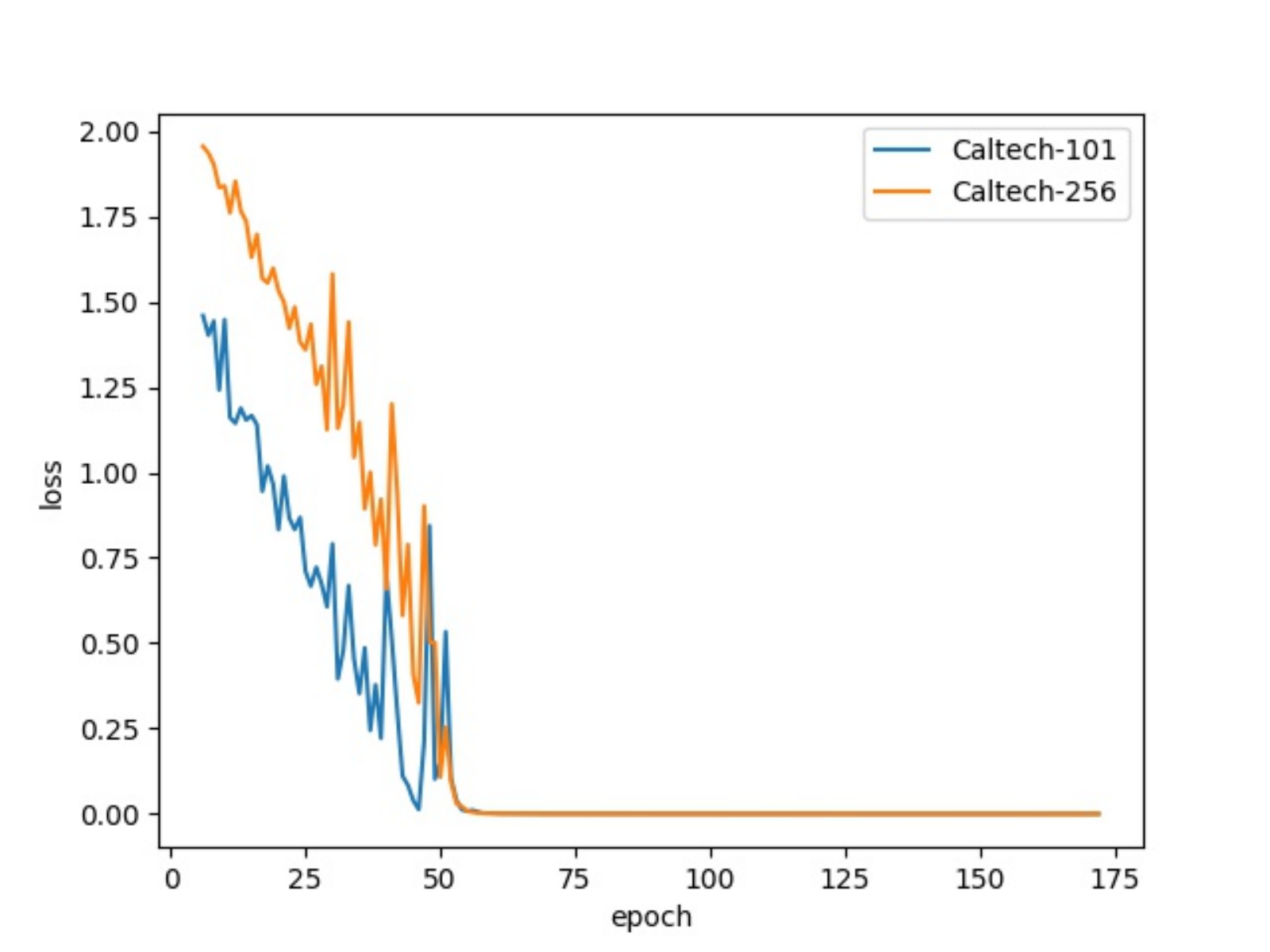}
}
\subfigure[] {
\includegraphics[width=0.41\columnwidth]{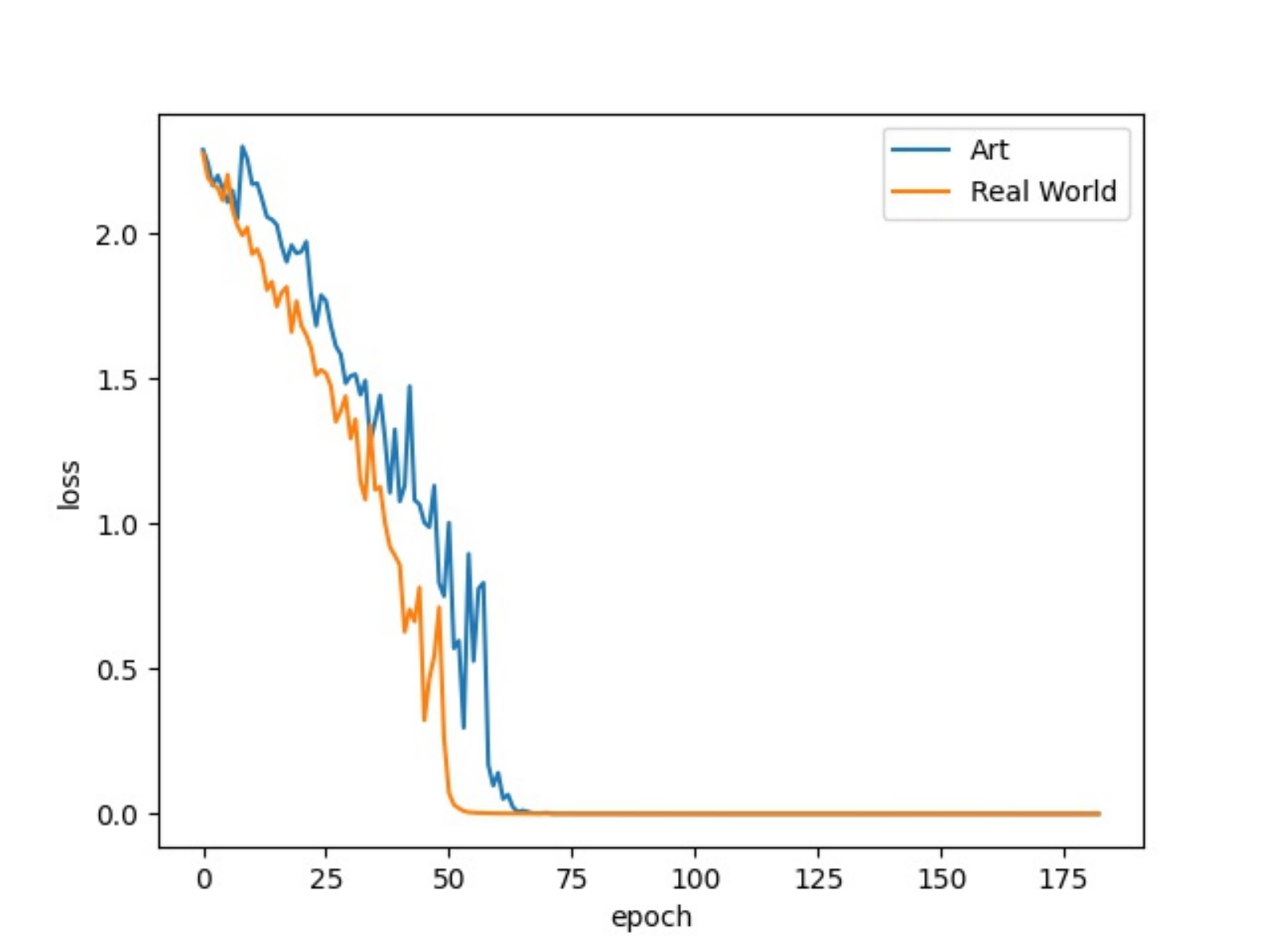}
}

\caption{Sub-Fig.(a) shows the convergence curve of DAMTL when the label sets among tasks partially overlapping; sub-fig. Sub-Fig.(b) shows the convergence curve of DAMTL when the label sets among network tasks do not overlap.}\label{Fig:hh}
\label{fig} \label{Fig:6}
\end{figure}

\section{Conclusion}
In this work, we provide a deep multi-task learning framework DAMTL, which is used to deal with multi-tasks with partial or unoverlapping label sets among tasks. Compared with the previous MTL method, DAMTL leverages big auxiliary tasks to jointly learn multiple tasks with partially overlapping or unoverlapping label sets. In addition, the auxiliary strategies in DAMTL can be flexibly embedded in other deep multi-task learning frameworks or transfer learning frameworks. In order to evaluate the performance of DAMTL, we conduct experiments on twelve public datasets and compared  state-of-the-art MTL methods. Experimental results show that the DAMTL framework has significant advantages. In summary, our work can enrich MTL research to a certain extent from two aspects: 1) a novel adaptive MT learning mechanism is used to deal with multiple tasks when the label sets are partially overlapped or even unoverlapped. 2) A new knowledge extraction strategy that uses a set of soft masking matrices to adaptively prune the hidden neurons in the auxiliary task network to extract specific knowledge that assist the current task learn to form a corresponding network for each task. However, in this work, we haven't solved the interpretable problems in MTL, and the following work will focus on such problems.

\section*{Acknowledgments}
This work is supported by NSFC under Grant No. 61672281, and the Key Program of NSFC under Grant No. 61732006. The authors would like to thank Dr. Meng Cao and Yingyu Zhong for their generous help and beneficial discussions.

\section*{References}

\bibliography{mybibfile}

\begin{thebibliography}{10}
\expandafter\ifx\csname url\endcsname\relax
  \def\url#1{\texttt{#1}}\fi
\expandafter\ifx\csname urlprefix\endcsname\relax\fi
\expandafter\ifx\csname href\endcsname\relax
  \def\href#1#2{#2} \def\path#1{#1}\fi

\bibitem{chen2018multi}
Y.~Chen, D.~Zhao, L.~Lv, Q.~Zhang, Multi-task learning for dangerous object
  detection in autonomous driving, Information Sciences 432 (2018) 559--571.

\bibitem{xu2020multi}
Q.~Xu, Y.~Zeng, W.~Tang, W.~Peng, T.~Xia, Z.~Li, F.~Teng, W.~Li, J.~Guo,
  Multi-task joint learning model for segmenting and classifying tongue images
  using a deep neural network, IEEE journal of biomedical and health
  informatics 24~(9) (2020) 2481--2489.

\bibitem{jiang2016novel}
Y.~Jiang, Z.~Deng, K.-S. Choi, F.-L. Chung, S.~Wang, A novel multi-task tsk
  fuzzy classifier and its enhanced version for labeling-risk-aware multi-task
  classification, Information Sciences 357 (2016) 39--60.

\bibitem{9053569}
M.~J.{Zhong}, S.~P.{Swietojanski}, J.{Monteiro}, J.{Trmal}, Y.{Bengio},
  Multi-task self-supervised learning for robust speech recognition, in: ICASSP
  2020-2020 IEEE International Conference on Acoustics, Speech and Signal
  Processing (ICASSP), 2020, pp. 6989--6993.
\newblock \href {http://dx.doi.org/10.1109/ICASSP40776.2020.9053569}
  {\path{doi:10.1109/ICASSP40776.2020.9053569}}.

\bibitem{ravanelli2020multi}
M.~Ravanelli, J.~Zhong, S.~Pascual, P.~Swietojanski, J.~Monteiro, J.~Trmal,
  Y.~Bengio, Multi-task self-supervised learning for robust speech recognition,
  in: ICASSP 2020-2020 IEEE International Conference on Acoustics, Speech and
  Signal Processing (ICASSP), IEEE, 2020, pp. 6989--6993.

\bibitem{zhao2020multi}
Z.~Zhao, J.~Qin, Z.~Gou, Y.~Zhang, Y.~Yang, Multi-task learning models for
  predicting active compounds, Journal of Biomedical Informatics 108 (2020)
  103484.

\bibitem{ruder2017sluice}
S.~Ruder, J.~Bingel, I.~Augenstein, A.~S{\o}gaard, Sluice networks: Learning
  what to share between loosely related tasks, stat 1050 (2017) 23.

\bibitem{2017arXiv170208303B}
J.~{Bingel}, A.~{S{\o}gaard}, {Identifying beneficial task relations for
  multi-task learning in deep neural networks}, arXiv e-prints (2017)
  arXiv:1702.08303\href {http://arxiv.org/abs/1702.08303}
  {\path{arXiv:1702.08303}}.

\bibitem{2016arXiv161101587H}
K.~{Hashimoto}, C.~{Xiong}, Y.~{Tsuruoka}, R.~{Socher}, {A Joint Many-Task
  Model: Growing a Neural Network for Multiple NLP Tasks}, arXiv e-prints
  (2016) arXiv:1611.01587\href {http://arxiv.org/abs/1611.01587}
  {\path{arXiv:1611.01587}}.

\bibitem{2015arXiv150602117L}
M.~{Long}, Z.~{Cao}, J.~{Wang}, P.~S. {Yu}, {Learning Multiple Tasks with
  Multilinear Relationship Networks}, arXiv e-prints (2015)
  arXiv:1506.02117\href {http://arxiv.org/abs/1506.02117}
  {\path{arXiv:1506.02117}}.

\bibitem{ma2018modeling}
J.~Ma, Z.~Zhao, X.~Yi, J.~Chen, L.~Hong, E.~H. Chi, Modeling task relationships
  in multi-task learning with multi-gate mixture-of-experts, in: Proceedings of
  the 24th ACM SIGKDD International Conference on Knowledge Discovery \& Data
  Mining, 2018, pp. 1930--1939.

\bibitem{2021arXiv210112431F}
Q.~{Feng}, S.~{Chen}, {Learning Twofold Heterogeneous Multi-Task by Sharing
  Similar Convolution Kernel Pairs}, arXiv e-prints (2021)
  arXiv:2101.12431\href {http://arxiv.org/abs/2101.12431}
  {\path{arXiv:2101.12431}}.

\bibitem{2020arXiv200500944W}
S.~{Wu}, H.~R. {Zhang}, C.~{R{\'e}}, {Understanding and Improving Information
  Transfer in Multi-Task Learning}, arXiv e-prints (2020) arXiv:2005.00944\href
  {http://arxiv.org/abs/2005.00944} {\path{arXiv:2005.00944}}.

\bibitem{evgeniou2004regularized}
T.~Evgeniou, M.~Pontil, Regularized multi--task learning, in: Proceedings of
  the tenth ACM SIGKDD international conference on Knowledge discovery and data
  mining, 2004, pp. 109--117.

\bibitem{honorio2010multi}
J.~Honorio, D.~Samaras, Multi-task learning of gaussian graphical models, in:
  ICML, 2010.

\bibitem{Liu2020MultiTaskDF}
Q.~Liu, X.~Li, Z.~He, N.~Fan, D.~Yuan, W.~Liu, Y.~Liang, Multi-task driven
  feature models for thermal infrared tracking, in: AAAI, 2020.

\bibitem{wang2021learning}
J.~Wang, S.~Zhang, Y.~Wang, Z.~Zhu, Learning efficient multi-task stereo
  matching network with richer feature information, Neurocomputing 421 (2021)
  151--160.

\bibitem{2020arXiv200204813G}
P.~{Guo}, C.~{Deng}, L.~{Xu}, X.~{Huang}, Y.~{Zhang}, {Deep Multi-Task
  Augmented Feature Learning via Hierarchical Graph Neural Network}, arXiv
  e-prints (2020) arXiv:2002.04813\href {http://arxiv.org/abs/2002.04813}
  {\path{arXiv:2002.04813}}.

\bibitem{2020arXiv200413379V}
S.~{Vandenhende}, S.~{Georgoulis}, W.~{Van Gansbeke}, M.~{Proesmans}, D.~{Dai},
  L.~{Van Gool}, {Multi-Task Learning for Dense Prediction Tasks: A Survey},
  arXiv e-prints (2020) arXiv:2004.13379\href {http://arxiv.org/abs/2004.13379}
  {\path{arXiv:2004.13379}}.

\bibitem{shen2020deep}
Z.~Shen, C.~Cui, J.~Huang, J.~Zong, M.~Chen, Y.~Yin, Deep adaptive feature
  aggregation in multi-task convolutional neural networks, in: Proceedings of
  the 29th ACM International Conference on Information \& Knowledge Management,
  2020, pp. 2213--2216.

\bibitem{2020Identifying}
S.~Yadav, J.~Chauhan, J.~P. Sain, K.~Thirunarayan, J.~Schumm, Identifying
  depressive symptoms from tweets: Figurative language enabled multitask
  learning framework, in: Proceedings of the 28th International Conference on
  Computational Linguistics, 2020.

\bibitem{2019arXiv191112423S}
X.~{Sun}, R.~{Panda}, R.~{Feris}, K.~{Saenko}, {AdaShare: Learning What To
  Share For Efficient Deep Multi-Task Learning}, arXiv e-prints (2019)
  arXiv:1911.12423\href {http://arxiv.org/abs/1911.12423}
  {\path{arXiv:1911.12423}}.

\bibitem{sun2020learning}
T.~Sun, Y.~Shao, X.~Li, P.~Liu, H.~Yan, X.~Qiu, X.~Huang, Learning sparse
  sharing architectures for multiple tasks, in: Proceedings of the AAAI
  Conference on Artificial Intelligence, Vol.~34, 2020, pp. 8936--8943.

\bibitem{2020arXiv200811643V}
S.~{Verboven}, M.~{Hafeez Chaudhary}, J.~{Berrevoets}, W.~{Verbeke},
  {HydaLearn: Highly Dynamic Task Weighting for Multi-task Learning with
  Auxiliary Tasks}, arXiv e-prints (2020) arXiv:2008.11643\href
  {http://arxiv.org/abs/2008.11643} {\path{arXiv:2008.11643}}.

\bibitem{sanh2019hierarchical}
V.~Sanh, T.~Wolf, S.~Ruder, A hierarchical multi-task approach for learning
  embeddings from semantic tasks, in: Proceedings of the AAAI Conference on
  Artificial Intelligence, Vol.~33, 2019, pp. 6949--6956.

\bibitem{baxter1997bayesian}
J.~Baxter, A bayesian/information theoretic model of learning to learn via
  multiple task sampling, Machine learning 28~(1) (1997) 7--39.

\bibitem{ruder2019latent}
S.~Ruder, J.~Bingel, I.~Augenstein, A.~S{\o}gaard, Latent multi-task
  architecture learning, in: Proceedings of the AAAI Conference on Artificial
  Intelligence, Vol.~33, 2019, pp. 4822--4829.

\bibitem{Strezoski_2019_ICCV}
G.~Strezoski, N.~v. Noord, M.~Worring, Many task learning with task routing,
  in: Proceedings of the IEEE/CVF International Conference on Computer Vision
  (ICCV), 2019.

\bibitem{fernando2017pathnet}
C.~Fernando, D.~Banarse, C.~Blundell, Y.~Zwols, D.~Ha, A.~A. Rusu, A.~Pritzel,
  D.~Wierstra, Pathnet: Evolution channels gradient descent in super neural
  networks, CoRR, abs/1701.0873,2017.

\bibitem{pironkov2020hybrid}
G.~Pironkov, S.~U. Wood, S.~Dupont, Hybrid-task learning for robust automatic
  speech recognition, Computer Speech \& Language (2020) 101103.

\bibitem{cao2017sparse}
P.~Cao, X.~Shan, D.~Zhao, M.~Huang, O.~Zaiane, Sparse shared structure based
  multi-task learning for mri based cognitive performance prediction of
  alzheimer’s disease, Pattern Recognition 72 (2017) 219--235.

\bibitem{2020arXiv200509910L}
S.~{Lee}, Y.~{Son}, {Multitask Learning with Single Gradient Step Update for
  Task Balancing}, arXiv e-prints (2020) arXiv:2005.09910\href
  {http://arxiv.org/abs/2005.09910} {\path{arXiv:2005.09910}}.

\bibitem{Sgaard2016DeepML}
A.~S{\o}gaard, Y.~Goldberg, Deep multi-task learning with low level tasks
  supervised at lower layers, in: ACL, 2016.

\bibitem{7849143}
J.~{Fan}, T.~{Zhao}, Z.~{Kuang}, Y.~{Zheng}, J.~{Zhang}, J.~{Yu}, J.~{Peng},
  Hd-mtl: Hierarchical deep multi-task learning for large-scale visual
  recognition, IEEE Transactions on Image Processing 26~(4) (2017) 1923--1938.
\newblock \href {http://dx.doi.org/10.1109/TIP.2017.2667405}
  {\path{doi:10.1109/TIP.2017.2667405}}.

\bibitem{xue2020weighted}
W.~Xue, Weighted feature-task-aware regularization learner for multitask
  learning, Pattern Analysis and Applications 23~(1) (2020) 253--263.

\bibitem{zhang2019multi}
J.~Zhang, J.~Miao, K.~Zhao, Y.~Tian, Multi-task feature selection with sparse
  regularization to extract common and task-specific features, Neurocomputing
  340 (2019) 76--89.

\bibitem{shao2020hypergraph}
W.~Shao, Y.~Peng, C.~Zu, M.~Wang, D.~Zhang, A.~D.~N. Initiative, et~al.,
  Hypergraph based multi-task feature selection for multimodal classification
  of alzheimer's disease, Computerized Medical Imaging and Graphics 80 (2020)
  101663.

\bibitem{2018Multi}
L.~Li, X.~Pan, H.~Yang, Z.~Liu, Y.~He, Z.~Li, Y.~Fan, Z.~Cao, L.~Zhang,
  Multi-task deep learning for fine-grained classification and grading in
  breast cancer histopathological images, Multimedia (2020) 14509--15428.

\bibitem{zheng2019metadata}
Z.~Zheng, Y.~Wang, Q.~Dai, H.~Zheng, D.~Wang, Metadata-driven task relation
  discovery for multi-task learning., in: IJCAI, 2019, pp. 4426--4432.

\bibitem{9098703}
C.~{Yan}, J.~{Xu}, J.~{Xie}, C.~{Cai}, H.~{Lu}, Prior-aware cnn with multi-task
  learning for colon images analysis, in: 2020 IEEE 17th International
  Symposium on Biomedical Imaging (ISBI), 2020, pp. 254--257.
\newblock \href {http://dx.doi.org/10.1109/ISBI45749.2020.9098703}
  {\path{doi:10.1109/ISBI45749.2020.9098703}}.

\bibitem{misra2016cross}
I.~Misra, A.~Shrivastava, A.~Gupta, M.~Hebert, Cross-stitch networks for
  multi-task learning, in: Proceedings of the IEEE Conference on Computer
  Vision and Pattern Recognition, 2016, pp. 3994--4003.

\bibitem{duan2020unsupervised}
R.~Duan, N.~F. Chen, Unsupervised feature adaptation using adversarial
  multi-task training for automatic evaluation of children’s speech, Proc.
  Interspeech (2020) 3037--3041.

\bibitem{2017arXiv170400514A}
I.~{Augenstein}, A.~{S{\o}gaard}, {Multi-Task Learning of Keyphrase Boundary
  Classification}, arXiv e-prints (2017) arXiv:1704.00514\href
  {http://arxiv.org/abs/1704.00514} {\path{arXiv:1704.00514}}.

\bibitem{rai2010infinite}
P.~Rai, H.~Daum{\'e}~III, Infinite predictor subspace models for multitask
  learning, in: Proceedings of the Thirteenth International Conference on
  Artificial Intelligence and Statistics, 2010, pp. 613--620.

\bibitem{Zhou_2020_CVPR}
L.~Zhou, Z.~Cui, C.~Xu, Z.~Zhang, C.~Wang, T.~Zhang, J.~Yang, Pattern-structure
  diffusion for multi-task learning, in: Proceedings of the IEEE/CVF Conference
  on Computer Vision and Pattern Recognition (CVPR), 2020.

\bibitem{article}
Y.~Wang, X.~Luo, L.~Ding, S.~Fu, S.~Hu, Multi-task non-negative matrix
  factorization for visual object tracking, Pattern Analysis and Applications
  2020.
\newblock \href {http://dx.doi.org/10.1007/s10044-019-00812-4}
  {\path{doi:10.1007/s10044-019-00812-4}}.

\bibitem{8944708}
J.~{Jeong}, C.~{Jun}, Sparse tensor decomposition for multi-task interaction
  selection, in: 2019 IEEE International Conference on Big Knowledge (ICBK),
  2019, pp. 105--114.
\newblock \href {http://dx.doi.org/10.1109/ICBK.2019.00022}
  {\path{doi:10.1109/ICBK.2019.00022}}.

\bibitem{2020Predicting}
F.~Huang, Y.~Qiu, Q.~Li, S.~Liu, F.~Ni, Predicting drug-disease associations
  via multi-task learning based on collective matrix factorization, Frontiers
  in Bioengineering and Biotechnology 8 (2020) 218.

\bibitem{2020arXiv200204799Z}
Y.~{Zhang}, Y.~{Zhang}, W.~{Wang}, {Deep Multi-Task Learning via Generalized
  Tensor Trace Norm}, arXiv e-prints (2020) arXiv:2002.04799\href
  {http://arxiv.org/abs/2002.04799} {\path{arXiv:2002.04799}}.

\bibitem{chen2020template}
Z.~Chen, H.~Lei, Y.~Zhao, Z.~Huang, X.~Xiao, Y.~Lei, E.-L. Tan, B.~Lei,
  Template-oriented multi-task sparse low-rank learning for parkinson’s
  diseases diagnosis, in: International Workshop on PRedictive Intelligence In
  MEdicine, Springer, 2020, pp. 178--187.

\bibitem{8694882}
X.~{Wu}, X.~{Zhang}, Y.~{Cen}, Multi-task joint sparse and low-rank
  representation target detection for hyperspectral image, IEEE Geoscience and
  Remote Sensing Letters 16~(11) (2019) 1756--1760.
\newblock \href {http://dx.doi.org/10.1109/LGRS.2019.2908196}
  {\path{doi:10.1109/LGRS.2019.2908196}}.

\bibitem{long2014transfer}
M.~Long, J.~Wang, G.~Ding, J.~Sun, P.~S. Yu, Transfer joint matching for
  unsupervised domain adaptation, in: Proceedings of the IEEE conference on
  computer vision and pattern recognition, 2014, pp. 1410--1417.

\bibitem{2014arXiv1409.1556S}
K.~{Simonyan}, A.~{Zisserman}, {Very Deep Convolutional Networks for
  Large-Scale Image Recognition}, arXiv e-prints (2014) arXiv:1409.1556\href
  {http://arxiv.org/abs/1409.1556} {\path{arXiv:1409.1556}}.

\bibitem{misra2016cross-stitch}
I.~Misra, A.~Shrivastava, A.~Gupta, M.~Hebert, Cross-stitch networks for
  multi-task learning (2016) 3994--4003.

\bibitem{gao2019nddr-cnn:}
Y.~Gao, J.~Ma, M.~Zhao, W.~Liu, A.~L. Yuille, Nddr-cnn: Layerwise feature
  fusing in multi-task cnns by neural discriminative dimensionality reduction
  (2019) 3205--3214.

\end{thebibliography}

\end{document}